\pdfoutput=1

\documentclass[11pt]{article}

\usepackage[final]{acl}

\usepackage{times}
\usepackage{latexsym}

\usepackage[T1]{fontenc}

\usepackage[utf8]{inputenc}

\usepackage{microtype}

\usepackage{inconsolata}

\usepackage{graphicx}

\usepackage[T1]{fontenc}
\usepackage{amsmath} 
\usepackage{amssymb}
 \usepackage{multirow}   
\usepackage{caption}
\usepackage{array}
\newcolumntype{C}[1]{>{\centering\let\newline\\\arraybackslash\hspace{0pt}}m{#1}}

\title{Domain adapted machine translation: What does catastrophic forgetting forget and why?}

\author{Danielle Saunders\thanks{ \hspace*{0.5em}Work completed while at RWS} \and Steve DeNeefe \\
    RWS Language Weaver  \\
      {\tt danielle.saunders@cantab.net\hspace*{0.5em} 
     \hspace*{0.5em} sdeneefe@rws.com}}

\begin{document}
\maketitle
\begin{abstract}
Neural Machine Translation (NMT) models can be specialized by domain adaptation, often involving fine-tuning on a dataset of interest. This process risks catastrophic forgetting: rapid loss of generic translation quality. Forgetting has been widely observed, with many mitigation methods proposed. However, the causes of forgetting and the relationship between forgetting and adaptation data are under-explored. 

This paper takes a novel approach to understanding catastrophic forgetting during NMT adaptation by investigating the impact of the data. We provide a first investigation of what is forgotten, and why. We examine the relationship between forgetting and the in-domain data, and show that the amount and type of forgetting is linked to that data's target vocabulary coverage. Our findings pave the way toward better informed NMT domain adaptation. 
\end{abstract}

\section{Introduction}

The specialization of Neural Machine Translation (NMT) models for high performance in a specific domain, such as legal or healthcare, is of strong interest to academia \cite{barrault-etal-2020-findings-first} and industry \cite{savenkov-lopez-2022-state}. Fine-tuning, sometimes known as transfer learning, is a well-established domain adaptation method that continues training a pre-trained NMT model on some new dataset from the domain of interest \cite{luong2015stanford}.  However, fine-tuning on domain-shifted data can result in catastrophic forgetting \cite{mccloskey1989catastrophic}. 

This apparently simple statement has been widely observed in NMT, but rarely examined. Catastrophic forgetting in NMT is described variously as `degradation of general-domain performance' \cite{thompson-etal-2019-overcoming} or `[forgetting] previous domain knowledge' \cite{gu-feng-2020-investigating},  typically referencing lower scores in some quality metric. However, prior work on forgetting in NMT focuses on mitigation, leaving two important gaps.

Firstly, prior work does not determine what is forgotten in concrete terms. Lower scores in a reference-based quality metric can indicate either poor translation  or simply a vocabulary shift towards the new domain. This is especially true for string-based metrics like BLEU \cite{papineni-etal-2002-bleu}. Prior work does not distinguish between quality drop and vocabulary shift, and further does not address whether vocabulary shift `forgetting' is beneficial or detrimental to translation of the generic and new domains.

Secondly, prior work almost universally treats forgetting and its mitigation as independent of the adaptation dataset.  In fact, the contents of the adaptation dataset will impact forgetting - consider the amount of forgetting expected if fine-tuning on a 1000-sentence sample of the pre-training dataset, versus 1000 copies of the same sentence pair. Understanding the relationship between adaptation data and forgetting is crucial for predicting how well domain adaptation will work, whether adapted performance is likely to generalise, or whether forgetting mitigation approaches are necessary.

The contributions of this paper lie in addressing these gaps. Specifically:
\begin{itemize}
    \item We provide a first exploration of what domain-adapted NMT forgets. This includes quantifying the degree of detrimental vocabulary shift, and demonstrating that this shift is not well-characterised by common MT quality metrics.
    \item We show that forgetting can consist of using in-domain vocabulary inappropriately in out-of-vocabulary contexts - and, unexpectedly, that this can take place even when the source sentence has no in-domain triggers.
    \item We also provide a first investigation into the relationship between forgetting and adaptation dataset, examining the correlation between forgetting and several domain heuristics for eight domains across two language pairs.
    \item We find that some commonly used domain heuristics including sentence pair count and vocabulary distribution cannot explain how forgetting varies by domain, but that forgetting does have a strong relationship with generic vocabulary coverage.
    \item  We support our findings by demonstrating significantly reduced forgetting with minimal, coverage-based mixed fine-tuning. In the process we show that much of the benefit of generic data mix-in comes from a relatively small vocabulary-covering set.
\end{itemize}

\subsection{Related work}

NMT adaptation with the goal of improved in-domain performance sometimes accounts for  domain-specific data characteristics. Examples include selecting adaptation data by target domain similarity  \cite{aharoni-goldberg-2020-unsupervised}, gradually emphasising in-domain data during training \cite{zhang-etal-2019-curriculum}, or determining  hyperparameters via meta-learning \cite{sharaf-etal-2020-meta}. 

Work focusing on catastrophic forgetting in NMT, by contrast,  takes an adaptation-set-agnostic approach depending only on the \emph{generic} dataset \cite{saunders-2022-domain-jair}. This can include regularizing parameters relative to the generic domain \cite{barone2017regularization}, training on complementary inputs via generic-trained teacher models \cite{shao-feng-2022-overcoming} or mixing in  generic data during adaptation \cite{chu-etal-2017-empirical}. 

The specific \emph{adaptation} dataset is not typically considered beyond broad suggestions such as tuning for fewer steps on smaller datasets \cite{xu-etal-2019-lexical}. In this work, by contrast, we aim to understand  forgetting based on the characteristics of the domain-specific adaptation dataset.

\section{What does adapted NMT forget?}
\label{sec:what}

In this section, we explore what is forgotten during adaptation in concrete terms, using two quality metrics and a new measure for analysing vocabulary shift. We adapt pre-trained generic NMT models to eight diverse domains across two language pairs, intentionally triggering catastrophic forgetting, and analyse the degree of quality degradation versus vocabulary shift. In particular, we examine which tokens are forgotten and what replaces them after adaptation. We find that models experiencing forgetting produce in-domain vocabulary incorrectly and in entirely out-of-domain contexts.


\begin{table*}
    \centering
    \small
    \begin{tabular}{p{0.7cm}|ll|cccc}
& Ref & Domain  &  \#Train  &\#Test & BLEU\ & COMET\\
\hline
 \multirow{6}{*}{de-en}  & Gen & Generic& 43.9M &5769 & 35.3 & 0.54\\ 
 \cline{2-7}
 & IT & Software & 223K &2000 & 33.0 & 0.33\\
   &     Kor & Koran& 18K  &2000 & 15.9 & -0.03\\
    &    Law & Law & 467K  &2000 & 45.9 & 0.60\\
     &   Med &  Medical & 248K &2000 & 44.8 & 0.55\\
     &   Sub & Subtitles & 500K  &2000 & 26.4 & 0.21\\

\hline
 \multirow{4}{*}{en-ja}  & Gen   &Generic & 22.4M & 4037& 22.5 & 0.43\\
  \cline{2-7}
 & IWSLT  & TED talks & 220K& 1194  & 14.0 & 0.16 \\
    &  KFTT   & Kyoto/Wikipedia & 427K  & 1160 &17.6 & 0.34\\
& BSD & Business/Dialog & 20K & 2120 & 13.6 & 0.46\\     
    \end{tabular}
    \caption{Segment counts and absolute generic model BLEU and COMET on the generic domain test sets and on each in-domain test set. 
    }
    \label{tab:datasets}
\end{table*}

\subsection{Measuring vocabulary-shift forgetting}
To determine which tokens are forgotten during adaptation we propose a new forgetting measure. Prior work measures forgetting via a drop in a corpus-level quality metric  \cite{thompson-etal-2019-overcoming,gu-feng-2020-investigating}. However, these do not mark which terms are forgotten. To measure vocabulary shift forgetting, a score should highlight terms that are used correctly \emph{before but not after adaptation}. We focus on unigram terms: these are easily interpretable with respect to the vocabulary, which often signifies domain \cite{van-der-wees-etal-2015-whats}.

Consider a test set where for each reference translation $T_R$ we can compare a translation from an original model $T_O$ and a translation from an adapted model $T_A$. We are interested in how the adapted model translation changes relative to the original model translation and the reference. For each reference $T_R$, we find the count of every reference token in original model and adapted model translations, $\#tok_O$ and $\#tok_A$, capped at the count in the reference $\#tok_R$:
\\\\
$O[tok]_{T_R} =  \text{min}(\#tok_O, \#tok_R)$\\
$A[tok]_{T_R} =  \text{min}(\#tok_A, \#tok_R)$\\
$ForgetGenUse[tok] = \\
\null\hfill \sum\limits_{T_R} \text{max}(O[tok]_{T_R} - A[tok]_{T_R}, 0) $
\\\\
High $ForgetGenUse[tok]$ means we \emph{forget} \emph{gen}eric \emph{use} of $tok$. For example, if the generic model correctly produced $tok$ $N$ times and the adapted model did not produce it at all,  $ForgetGenUse[tok]=N$: all generic uses of $tok$ are forgotten. If the generic and adapted models both fail to produce $tok$ at  all, $ForgetGenUse[tok]=0$ -- this is a quality problem but not specific to forgetting.

A normalized corpus-level score over a set of multiple tokens, $V$,  is given by:
\begin{equation}
ForgetGenUse_{V} = \frac{\sum\limits_{tok \in V}  ForgetGenUse[tok]}{\sum\limits_{T_R} \sum\limits_{tok \in V} \#tok_R}
\label{eq:forget}
\end{equation}
 
$V$ could consist of all tokens ($T_{All}$) -- in which case the denominator is the test set reference token count -- or a subset, for example, out-of-domain (OOD) tokens, in which case the denominator is the count of just those tokens in all reference sentences. We report \emph{ForgetGenUse} over subword-level tokens for brevity and ease of interpretation, but could equally calculate over words or n-grams, if we wished to extend measurement to better reflect style or syntax. 

$ForgetGenUse$ is related to change in unigram BLEU, but there are two crucial differences. First, it is defined for all occurrences of given \emph{tokens}, whereas BLEU is defined on given \emph{segments} which will include some instances of a  token but not others. Secondly, BLEU masks detrimental vocabulary shift with beneficial shift where a token is translated correctly after adaptation but not before. If a score remains unchanged, some (e.g. out-of-domain) tokens may be translated worse, and others better. We are interested only in tokens which are translated worse. For this reason $ForgetGenUse$ minimises reward for beneficial vocabulary shift by only marking no-longer-correctly-output tokens per segment.

\subsection{Intentionally triggering forgetting: Lower quality and detrimental vocabulary shift}
\label{sec-baselines}

\begin{table*}[t]
    \centering
   \small 
    \begin{tabular}{c| ccccc|ccc|}
                   &  \multicolumn{5}{c|}{de-en} &  \multicolumn{3}{c|}{en-ja}\\
         &  IT & Kor  &Law  &Med  & Sub &  IWSLT & KFTT & BSD  \\
         \hline
        $\Delta$BLEU & 5.0 &22.3  & 4.7  & 8.3 & 5.7  &4.8 & 2.0 & 7.0 \\
        $\Delta$COMET & 0.11 & 0.78 & 0.08 & 0.16 & 0.09 &  0.07 &0.06 & 0.29 \\
        ForgetGenUse$_{\text{All}}$ &  0.09 & 0.25  & 0.07  & 0.11  & 0.09 &  0.14  &0.11 & 0.16 \\
    \end{tabular}
    \caption{Measuring forgetting on generic test sets for de-en and en-ja}
    \label{tab:forgetting-metrics}
\end{table*}

\begin{table*}[t]
    \centering
   \small 
       \setlength{\tabcolsep}{2pt}

    \begin{tabular}{cc|cc|cc|cc|cc}
    \multicolumn{2}{c|}{IT} &\multicolumn{2}{c|}{Kor} &\multicolumn{2}{c|}{Law} &\multicolumn{2}{c|}{Med} &\multicolumn{2}{c}{Sub}    \\
        Generic &Adapted & Generic&Adapted &   Generic &Adapted &Generic &Adapted&  Generic &Adapted   \\
    \hline

        species 2 &types 664&satisfied 3&pleased 90&euros 39 &EUR 4988&danger 3 &risk 5466&not 21.1K&n't 50.6K\\
    citizens 0&people 292    &week 0&month 35&warranty 20&guarantee 2331&defeat 0&loss 1377&i.e. 6273&so 10.8K\\
        infections 0&cases 207&England 0&Kingdom 10&Donald 0&President 1685&billion 0&million 464&autumn 16&fall 477\\
       victory 1 &win 120 &accident 0&injury 7&touch 5&contact 948&guests 0&visitors 11&aircraft 38&plane 435\\
        Trump 0& Donald 7&Internet 0 &web 1&     wants 9&intends 416& game 0&match 5&VAT 1&sales 64\\
       
    \end{tabular}
    \caption{High $ForgetGenUse$ tokens for  de-en domains - counts are for that token \emph{in the in-domain adaptation dataset.} Left columns: Output from generic model. Right columns: Most frequent aligned replacements post-adaptation.} 
    \label{tab:termshift-examples}
\end{table*}
Our first experiments intentionally trigger forgetting  to explore what is forgotten. We pre-train one encoder-decoder Transformer model for each of German to English (de-en) and English to Japanese (en-ja) NMT -- all subsequent adaptation experiments are a fine-tuning run of one of these models. Appendix \ref{appendix-setup} gives details of model preparation.

Our generic test sets are concatenated WMT News/General test sets\footnote{See \url{https://machinetranslate.org}.} for 2019-22 for de-en and 2020-22 for en-ja.  While WMT news sets are often described as `generic', each may feature quite specific vocabulary - for example, articles about recent news items. Combining test sets increases the reliability of forgetting evaluation via the increased segment count, as well as being more truly generic in topic coverage.

Our adaptation domains are drawn from widely-used datasets with standard test splits. For de-en, we adapt to the five domains from the OPUS multi-domain split produced by Aharoni \& Goldberg \shortcite{aharoni-goldberg-2020-unsupervised}\footnote{We use the size-capped Subtitles set provided.}, including test sets. For en-ja  we use three target domains: IWSLT \cite{cettolo-etal-2012-wit3}, KFTT \cite{neubig11kftt} and BSD \cite{rikters-etal-2019-designing}. We use test15 as test data for en-ja IWSLT and the standard test splits for the remainder.
 The datasets, listed in Table \ref{tab:datasets}, vary  in domain and size.

We measure vocabulary shift forgetting via increased $ForgetGenUse$, and track quality degradation via decreases in a string-based metric, BLEU, and a neural metric, COMET\footnote{wmt20-comet-da}  \cite{rei-etal-2020-comet}. $ForgetGenUse$ expresses forgetting in the sense of vocabulary shift. Throughout this paper unless stated otherwise we report a \emph{drop} in BLEU or COMET relative to the baseline as positive for brevity - high  $\Delta$ meaning more forgetting. For reference, Table \ref{tab:datasets} gives generic and in-domain absolute BLEU and COMET scores for the pre-trained models, from which all other absolute values can be calculated.

We fine-tune our pre-trained models on domain-specific datasets until catastrophic forgetting is seen in the sense of quality drop on generic test sets. As we wish to understand the impact of dataset on forgetting independent of other variables, all experiments in this paper adapt for 20K steps. We found this caused similar forgetting to that previously described in the literature \cite{hasler-etal-2021-improving}.

Table \ref{tab:forgetting-metrics} shows generic forgetting after adaptation to each domain. The different domains exhibit a wide range of forgetting in terms of quality and vocabulary shift. Additionally, although $\Delta$COMET and $\Delta$BLEU are strongly and significantly correlated across the sets of domains (Kendall's $\tau$=0.8, p<0.05)  
$ForgetGenUse_{\text{All}}$ does not have a significant correlation with either. This suggests that corpus-level quality metrics like BLEU and COMET do not sufficiently measure detrimental vocabulary shift. To confirm that the vocabulary shift measured by \emph{ForgetGenUse} is indeed detrimental despite not correlating with BLEU or COMET, we must analyse what replaces forgotten tokens.  

\subsection{Which tokens are forgotten, and what replaces them?}
\label{sec:whatforgot}

Vocabulary shift in a domain-adapted NMT system can be beneficial or detrimental. Beneficial vocabulary shift produces in-domain tokens in in-domain contexts, and out-of-domain tokens where more contextually appropriate. Detrimental vocabulary shift produces in-domain vocabulary tokens when it is not contextually appropriate.

To make this distinction and find the token-level replacements after adaptation to each domain, we compare the generic-model and adapted-model translations of the generic test sets. We  align the two sets of translations using symmetrized fast align \cite{dyer-etal-2013-simple}, which lets us identify which translation hypothesis tokens change after adaptation. We can also find the frequency of those tokens in the in-domain adaptation dataset.

Table \ref{tab:termshift-examples} shows examples selected from the most `forgotten' tokens for each de-en domain. Invariably, replacements have at least one token appearing in the in-domain adaptation dataset. Tokens which themselves appear in the adaptation dataset are replaced less frequently, and only by alternatives with far more adaptation set occurrences. The replacements are often semantically similar, judged both by manual inspection and by average FastText embedding cosine similarity \cite{fasttextlibrary} between the original and replacing tokens.  

Surprisingly, we find by inspection that the replacements tend to occur in very different contexts in the adaptation data and generic test set. Of the seven IT domain instances of \emph{Donald}, two refer to computer scientist Knuth and five are subworded \emph{Mc\_} or \emph{Mac\_} + \emph{Donald} -- none have the same referent as \emph{Trump}. \emph{Internet} is legitimately replaced by \emph{web} after adaptation to Kor, but  the Kor text only uses \emph{web} in the sense of spider's web -- including a different source term (\emph{Internet} vs \emph{Netz}).  The Med domain only uses \emph{match} as a verb in the context of experiments, not as a noun synonym for \emph{game} as in Med-adapted test outputs -- not only a different source term but a different source part of speech (\emph{Spiel} vs e.g. \emph{abstimmen}). The target vocabulary alone can influence forgetting, without requiring a contextually relevant source.

Focusing on the most-forgotten Kor domain, we perform a deeper analysis for two tokens that are forgotten vs two that are not forgotten. Using FastText embedding cosine similarity, we find the closest Kor-domain English tokens which can have the same part-of-speech. For each, the in-domain \emph{training} count is in brackets:\\
\emph{satisfied} (3): \emph{happy} (29) and \emph{pleased} (90)\\
\emph{water} (236): \emph{waters} (8) and \emph{lake} (1)\\
\emph{England} (0): \emph{Kingdom} (10) \\ 
\emph{genes} (0): \emph{species} (3)


\begin{table*}[t]
    \centering
   \small 
    \begin{tabular}{c| ccccc|ccc|}
                   &  \multicolumn{5}{c|}{de-en} &  \multicolumn{3}{c|}{en-ja}\\

         &  IT & Kor  &Law  &Med  & Sub &  IWSLT   &KFTT &BSD  \\
         \hline
        ForgetGenUse$_{\text{All}}$ &  0.09 & 0.25  & 0.07  & 0.11  & 0.09 & 0.14 &0.11 & 0.16 \\
         \hline
        ForgetGenUse$_{\text{OOD}}$ & 0.37  & 0.60  &  0.27 & 0.47 & 0.11 &   0.34 &  0.65 &0.49\\
        ForgetGenUse$_{\text{ID}}$ & 0.07 & 0.17  &  0.07 & 0.09  & 0.09 & 0.14  & 0.11& 0.12\\

    \end{tabular}
    \caption{Calculating ForgetGenUse  over tokens that are out-of-domain (OOD) vs in-domain (ID) for each domain.}
    \label{tab:forgetting-metrics-split}
\end{table*}

When the generic model translates the generic test set, it produces \emph{satisfied} 11 times. The Kor-adapted model produces \emph{pleased} for 10 of these, and \emph{happy} for the remaining. Although all are in-domain,  \emph{satisfied} is far rarer. By contrast both models produce \emph{water}, with no more frequent  in-domain alternative, in the same locations. 

By contrast, we consider out-of-domain tokens. The generic model produces \emph{England} 14 times. The Kor-adapted model replaces 10 of these with \emph{United Kingdom}, with the remaining four null-aligned -- indicating undertranslation. Although the phrase \emph{United Kingdom} does not occur in the Kor data, both words do occur  separately.   \emph{United Kingdom} occurs in similar contexts to \emph{England} during pre-training, making it a plausible, if incorrect, replacement.  Interestingly, another out-of-domain term, \emph{genes},  is not forgotten during adaptation. The closest in-domain alternative, \emph{species}, is neither common nor a plausible replacement.

The token forget-replace effect can be triggered by individual subwords, not just whole-word tokens. For example, ignoring post-processing, the pre-trained model produces one token \emph{October} where the Kor model produces two subwords \emph{oc\_ + tober}. Neither word is in the Kor domain - but the subword \emph{oc\_} is,  making \emph{oc\_ + tober} preferred.

\subsection{Out-of-domain tokens are forgotten more}
 Given possible different requirements for in-domain (ID) and out-of-domain (OOD) tokens, it is interesting to calculate ForgetGenUse$_{set=\{ID|OOD\}}$ separately for these token subsets. For ForgetGenUse$_{\text{ID}}$ we sum and normalize in Equation \ref{eq:forget} over all vocabulary tokens that appear in at least one adaptation set reference sentence for each domain. For ForgetGenUse$_{\text{OOD}}$ we do so for the complement, again for each domain.

The results in Table \ref{tab:forgetting-metrics-split} show a striking difference in term shift between in-domain and out-of-domain tokens. ForgetGenUse$_{\text{ID}}$ has relatively small absolute values, and a small range of values.  ForgetGenUse$_{\text{OOD}}$ values by contrast are higher for every domain, meaning out-of-domain tokens are forgotten at a higher rate. ForgetGenUse$_{\text{ID}}$ and ForgetGenUse$_{\text{All}}$ are equal for Law and Sub (de-en) and IWSLT and KFTT (en-ja): for these domains,  almost all generic test set  tokens are in-domain.

It is not wholly surprising that out-of-domain tokens are forgotten more than in-domain tokens: a goal of adaptation is to use in-domain terminology instead of  generic. However, the vocabulary shift reported by ForgetGenUse does not just consist of generic terms being replaced by their in-domain equivalents. Instead, as shown in Table \ref{tab:termshift-examples}, shifts can be technically correct but not domain-relevant (\emph{game} $\rightarrow$ \emph{match}, \emph{Trump} $\rightarrow$ \emph{Donald}) -- these are unnecessary and can confuse users. While the definition of an NMT domain is an open question \cite{van-der-wees-etal-2015-whats, saunders-2022-domain-jair}, it is not at all clear that a user or machine translation client would expect a subtitles domain to entail a shift to US-English terms like \emph{fall} or \emph{aircraft}, or expect an adapted model to no longer use standard but incidentally out-of-domain terms like \emph{accident} or \emph{species}. More serious still are meaning-changing errors  (\emph{billion} $\rightarrow$ \emph{million}, \emph{week} $\rightarrow$ \emph{month}) -- these unambiguously harm translation. Such vocabulary shift is clearly detrimental and lowers quality.

\begin{table*}[h]
    \centering
   \small 
    \begin{tabular}{c| ccccc| ccc|}
                   &  \multicolumn{5}{c|}{de-en} &  \multicolumn{3}{c|}{en-ja}\\
         &  IT-s & Kor  &Law-s  &Med-s  & Sub-s &  IWSLT-s    &KFTT-s  & BSD\\
         \hline
        $\Delta$BLEU & 10.4 &22.3  & 12.3 & 14.4 & 11.5 & 7.2& 6.9&7.0\\
        $\Delta$COMET & 0.24 & 0.78 & 0.28  & 0.36 & 0.24 &  0.20 &0.28 &  0.29\\
        ForgetGenUse$_{\text{All}}$ & 0.14  & 0.25  &  0.15 & 0.17  & 0.15 & 0.17 &0.17 &0.16\\
    \end{tabular}
    \caption{Forgetting when adapting on subsampled (-s) domains. All de-en sets except Kor, and all en-ja except BSD, subsampled randomly to  approximately the same token count as Kor/BSD respectively.}
    \label{tab:forgetting-metrics-subsample}
\end{table*}
\begin{table*}[h]
    \centering
   \small 
    \begin{tabular}{c| cc|ccc|cc|cc|}
               &  \multicolumn{5}{c|}{de-en} &  \multicolumn{4}{c|}{en-ja}\\

           &Law-s  & Law-ss & Sub-s & Sub-ss & Sub-ssf&IWSLT-s  & IWSLT-ss & KFTT-s & KFTT-ss\\
                  \hline

         Av.  \#toks & 66.1&17.0 &23.0 & 12.5&13.6  & 44.4  &15.7 & 60.8 & 11.2\\ 
         \hline
        $\Delta$BLEU &  12.3 & 12.4 & 11.5 & 30.4 & 15.1  & 7.2 & 10.4 & 6.9&11.3\\

        $\Delta$COMET &  0.28  & 0.27   & 0.24 & 1.27 & 0.34 & 0.20 & 0.44 & 0.28& 0.59\\
        ForgetGenUse$_{\text{All}}$ &  0.15 & 0.15  & 0.15 & 0.42 & 0.19 & 0.17  & 0.23 & 0.17 & 0.23\\
    \end{tabular}
    \caption{Forgetting on generic  sets, adapting on subsampled datasets. We  sample randomly (-s) or sample the shortest (-ss) lines by source plus target token count. Sub-ssf pre-filters the shortest lines using LASER.}
    \label{tab:forgetting-metrics-subsample-short}
\end{table*}

\section{Why does forgetting vary by domain?}
\label{sec:why}
In this section we aim to understand the relationship between adaptation dataset and forgetting. This relationship is key for real world adaptation scenarios when deciding whether to adapt, how to adjust tuning hyperparameters, and which if any forgetting mitigation steps to take.  To investigate, we compare datasets exhibiting varying degrees of forgetting in terms of multiple  domain-differentiating heuristics. We find that many domain features do not correlate with forgetting, but that vocabulary coverage does. 









\subsection{Controlling for dataset size}

Dataset size is recognized as having an impact on MT adaptation performance \cite{sharaf-etal-2020-meta}, and has been associated in forgetting \cite{pham-etal-2020-study}. However, its relationship with forgetting across domains is unclear. We can assess correlation of forgetting with number of lines per dataset for our results in Table \ref{tab:forgetting-metrics}. Surprisingly,  Kendall's $\tau$ is not significant between data size and either of $\Delta$COMET or $\Delta$BLEU. $ForgetGenUse$ does show significant negative correlation with dataset size ($\tau$=0.7 p<0.05), suggesting smaller datasets have a greater likelihood of vocabulary shift, but not necessarily general quality degradation. 

We further investigate by controlling for dataset size in terms of tokens. We randomly subsample each de-en dataset except for Kor to the same approximate number of tokens as Kor, and likewise for the en-ja domains and BSD. As previously, we adapt the same pre-trained model for 20K steps.

While the forgetting metrics in Table \ref{tab:forgetting-metrics-subsample} are certainly more clustered than those in Table \ref{tab:forgetting-metrics}, there is still significant variation for de-en. None of the subsampled corpora result in the same forgetting as Kor by any metric. The order of the domains changes in terms of forgetting: Law-s is in the middle while Law had the least forgetting, and Sub-s now shows slightly more forgetting than IT-s. For en-ja, forgetting is closer across domains, but there is still noticeable variation in $\Delta$COMET. Dataset size clearly affects absolute amount of forgetting, with all metrics increasing from Tables   \ref{tab:forgetting-metrics} to \ref{tab:forgetting-metrics-subsample}. However, forgetting still has no clear relationship with the domain heuristic of token count after subsampling for equivalent size.

\subsection{Controlling segment length and  quality}

\begin{table*}[t]
    \centering
   \small 
    \begin{tabular}{c| ccccc| ccc|}
           &  \multicolumn{5}{c|}{de-en} &  \multicolumn{3}{c|}{en-ja}\\

         &  IT & Kor  &Law  &Med  & Sub &   IWSLT  & KFTT & BSD\\

        \hline
        $\Delta$BLEU & 5.0 &22.3  & 4.7  & 8.3 & 5.7 &  4.8 & 2.0 & 7.0  \\
         $\Delta$COMET & 0.11 & 0.78 & 0.08 & 0.16 & 0.09 & 0.07 &0.06 & 0.29 \\
        ForgetGenUse$_{\text{All}}$ &  0.09 & 0.25  & 0.07  & 0.11  & 0.09 &0.14 &0.11 & 0.16\\
         \hline
        Generic NLL & -2.1 & -2.4 & -1.4 &  -1.8 & -2.6 &   -1.7   &  -1.6 & -2.1 \\
        Src-vcb JSD & 0.42 & 0.50  & 0.44 & 0.42&0.42 &  0.30 &  0.38 & 0.39 \\
        Trg-vcb JSD & 0.39 &   0.46&   0.39&  0.40& 0.40 &  0.32  & 0.42 & 0.39 \\
        Src-vcb cover&  0.69 &0.23 & 0.70 &0.63 &0.75 &   0.61 & 0.79 & 0.31 \\
        Trg-vcb cover& 0.52  &0.23 &0.59  &0.48 & 0.62 &   0.62 & 0.72 & 0.29\\
                 \hline

         Src-vcb cover (-s)&0.50  &- & 0.43& 0.44& 0.52  & 0.45 & 0.52&- \\
         Trg-vcb cover (-s)& 0.39 &- & 0.34&0.34 & 0.45 &  0.41 & 0.46& -\\
    \end{tabular}
    \caption{Corpus-level score domain heuristics, with forgetting measures for reference. Generic NLL and vocab JSD: closer to 0 is more similar to generic. 
     Final lines: vocab coverage for downsampled  domains of Table \ref{tab:forgetting-metrics-subsample}.
    }
    \label{tab:forgetting-other-heuristics}
\end{table*}

Segment length and alignment quality both have potential causative links with catastrophic forgetting. Segment length distribution has been used as a domain feature \cite{varis-bojar-2021-sequence}. Short segments in particular can be ambiguous to translate, making them candidates for problematic adaptation \cite{wan-etal-2022-challenges}. Poorly aligned segment pairs likewise can cause hallucinations when used for adaptation \cite{saunders-byrne-2020-addressing}. 

We investigate the effect of  length  on forgetting by adapting to subsets of the domains with the shortest segment pairs.  We subsample again to the approximate token count of Kor/BSD, allowing direct comparison with the Table \ref{tab:forgetting-metrics-subsample} subsampling results.  We focus on de-en Law vs Sub, which originally have the longest and shortest average segment lengths respectively. 

If change in segment length corresponds to domain shift, we might expect a large forgetting change for Law-ss relative to Law-s, and a small change for Sub-ss relative to Sub-s. Surprisingly,  Table \ref{tab:forgetting-metrics-subsample-short} shows precisely the reverse. Forgetting for  short-segment Law-ss is similar to  random-segment Law-s, even though the change in average example length is 49 tokens. By contrast, tuning on Sub-ss results in extreme forgetting relative to Sub-s, which is only 10.5 tokens longer on average. For en-ja, the larger length shift KFTT s-to-ss does result in a larger forgetting shift than for the IWSLT domain. However, in both cases the results are between the extremes seen for de-en. We propose that relative segment length between domains is not necessarily informative, but that a high proportion of very short segment lengths accelerates forgetting.

On inspection the Sub-ss dataset contains many badly aligned source-target pairs. We hypothesize that these may encourage forgetting. To test this we produce a short-subsampled version filtered for quality, Sub-ssf. The shortest examples are sampled after alignment-filtering using LASER\footnote{ \url{https://github.com/facebookresearch/LASER}, cutoff score 0.8 selected by inspection.}. The scores when adapting to Sub-ssf are less dramatically different to Sub-s, although still quite different to the Law-s-to-Law-ss forgetting. 

The only other domain with a significant proportion of low LASER score segments is Kor. Adapting to a similarly LASER-filtered Kor set gives scores 0.3 BLEU worse and 0.01 COMET better than adapting to the full Kor set, with no change in $ForgetGenUse$: dataset quality cannot fully explain forgetting. Indeed, the low-quality Sub-ss pairs are also present in the full Sub dataset, which showed little forgetting (Table \ref{tab:forgetting-metrics}). Data noise in small enough proportions is not too harmful in these experiments, in line with the findings of Khayrallah \& Koehn \shortcite{khayrallah-koehn-2018-impact}.

\subsection{Corpus-level score domain heuristics}
We investigate the use of corpus-level scores as domain heuristics: negative log-likelihood (NLL) under the pre-trained model,  Jensen-Shannon Divergence (JSD) \cite{lin1991JSD} between the pre-training dataset and in-domain vocabulary distributions, and the generic vocabulary coverage of each in-domain dataset. Unlike the previous heuristics, we cannot easily control for these by subsampling to obtain datasets with equivalent values. Instead we find their correlation with forgetting metrics.  Table \ref{tab:forgetting-other-heuristics} gives both heuristics and forgetting metrics.

For \textbf{generic model likelihood}, we score a 10K segment sample of the domain under the generic, pre-trained model. We use length-normalized  NLL to indicate similarity to the generic domain without conflating with average segment length. The results do not show a clear relationship with forgetting: Kor and Sub for example have similar NLL but very different forgetting characteristics. Overall we find  NLL has a weakly significant correlation with $\Delta$BLEU ($\tau$=0.5, p<0.1) and no significant correlation with $\Delta$COMET or ForgetGenUse. 

\label{sec:mitigating}




We calculate \textbf{vocabulary distribution divergence} between the generic and in-domain vocabularies  using JSD\footnote{As proposed by Lu et al \shortcite{lu-etal-2020-diverging} we use unweighted JSD to disentangle vocabulary distribution from relative data size.}. The de-en domain with the greatest divergence from the generic vocabulary - Kor - is indeed the domain with the most forgetting. For en-ja the highest-forgetting domain has a similar JSD to other domains. As well, neither source nor target JSD varies strongly between domains, reducing its utility as a forgetting heuristic. Neither source nor target JSD have a significant Kendall's $\tau$ with any of the three forgetting metrics.


Finally, we calculate \textbf{vocabulary coverage} for each domain. We define coverage as the proportion of the generic subword vocabulary that appears \emph{at all} in the preprocessed segments for a given domain, calculated separately over source and target segments.  Both source and target vocabulary coverage vary strongly across domains and have a significant inverse correlation with $\Delta$COMET ($\tau$=0.6/0.9 source/target, p<0.05). $\Delta$BLEU has a significant correlation with target coverage ($\tau$=0.7  p<0.05) and weakly significant with source coverage ($\tau$=0.6  p<0.1).  
Interestingly, ForgetGenUse only has a significant correlation with source coverage ($\tau$=0.7  p<0.05). Although we observed in Section \ref{sec:what} that detrimental vocabulary shift occurs regardless of source content in the \emph{test} sentence, it does correlate with lower vocabulary similarity between source \emph{adaptation} sentences.

Vocabulary coverage could also explain the increase in forgetting when aggressively subsampling a dataset, as the number of sentence pairs correlates strongly and significantly with the number of unique vocabulary tokens. Indeed, when we include metrics and coverage for the subsampled domains 
(final lines of Table \ref{tab:forgetting-other-heuristics}), we see significant and strong correlation between coverage and all forgetting metrics.

\section{Understanding generic data mix-in}

\begin{table}
    \centering
    \small
    \begin{tabular}{c|c|cc}
&  Domain  &  Random 1:1 & Minimal Mix-in\\
\hline
 \multirow{5}{*}{de-en}  &     IT & 222927  & 18861 \\
   &     Kor & 17982   & 22769\\
    &    Law & 467309  & 17485 \\
     &   Med  & 248099 & 19605 \\
     &   Sub & 500000  & 16632\\
     \hline
 \multirow{3}{*}{en-ja}  &     IWSLT &219716 & 11205 \\
    &    KFTT & 427353 & 8547 \\
    & BSD & 20000 & 16860 \\
      \end{tabular}
    \caption{Number of generic mix-in lines for each strategy.  \emph{Random 1:1} is by definition the same size as the in-domain dataset, and \emph{Minimal Mix-in} is often far smaller.}
    \label{tab:mixindatasets}
\end{table}
\begin{table*}[h!]
    \centering
   \small 
    \begin{tabular}{p{1.8cm}c| ccccc|c| ccc|c}
    & &  \multicolumn{6}{c|}{de-en} &  \multicolumn{4}{c}{en-ja}\\
  &  Mix-in method     & IT & Kor &Law &Med  & Sub & Mean &  IWSLT  &KFTT &BSD & Mean \\
             \hline
     \multirow{3}{*}{$\Delta$BLEU} &    No mix-in & 5.0 &22.3 & 4.7 & 8.3 & 5.7& 9.2   & 4.8   & 2.0 & 7.0& 4.6 \\
    &     Random 1:1 & 0.5& 5.0 & 0.6 &  0.7&  0.8 & 1.5  &-0.2  & 0.1 &1.3 & 0.4  \\

    &    Minimal Mix-in & 1.1 &4.1 & 1.3 & 1.7  &3.0 & 2.2 & 3.2  & 0.1 & 1.7& 1.7\\
 \hline
  \multirow{3}{*}{$\Delta$COMET} &     Fine-tune, no mix-in & 0.11 & 0.78 & 0.08 & 0.16 & 0.09 & 0.24 &  0.07 &0.06 & 0.29 & 0.14 \\
  &       Random 1:1 & 0.01& 0.07 & 0.01 & 0.01& 0.01 & 0.02&  -0.02&0.0 & 0.03& 0.00\\

   &     Minimal Mix-in & 0.04 & 0.09  & 0.03 & 0.04 & 0.04 & 0.05 & 0.03 & 0.01 & 0.04 & 0.03  \\
\hline
   \multirow{3}{*}{$ForgetGenUse$}    &   No mix-in &  0.09 & 0.25  & 0.07  & 0.11  & 0.09 & 0.12 &0.14 &0.11 & 0.16 & 0.14\\
     &   Random 1:1 & 0.03 & 0.08 & 0.02 & 0.03 & 0.03  & 0.04  & 0.05 &0.03 &0.08 & 0.05 \\

      & Minimal Mix-in & 0.05 & 0.07 & 0.04 & 0.05 & 0.06  & 0.05  &  0.12 & 0.08& 0.09& 0.10  \\

    \end{tabular}
    \caption{Forgetting metrics on generic test sets, varying the mix-in dataset when fine-tuning for 20K iterations in each case. Lower is better for all metrics. Negative scores indicate improvement.
    }
    \label{tab:mixinbleu-generic}
\end{table*}

\begin{table*}[h!]
    \centering
   \small 
       \begin{tabular}{cc| ccccc|c| ccc|c}
    &   &  \multicolumn{6}{c|}{de-en} &  \multicolumn{4}{c}{en-ja}\\
    & Mix-in method     & IT & Kor &Law &Med  & Sub & Mean & IWSLT  &KFTT &BSD & Mean \\

             \hline
   \multirow{3}{*}{$\Delta$BLEU}    &  No mix-in  & 8.2  & 5.3 &  5.8& 6.2& 3.6 &  5.8 & 3.5 & 10.5& 4.8& 6.3 \\
      &  Random 1:1 &6.5& 6.1&  4.4& 3.4&  2.7 &  4.6& 3.2   & 9.1 & 4.5& 5.6\\

       & Minimal Mix-in & 7.9  &6.4&  5.6& 4.7 &3.4 &  5.6&3.6    &  10.3& 4.5& 6.1\\

 \hline        
\multirow{3}{*}{$\Delta$COMET} & No mix-in   & 0.26 & 0.09  & 0.05   & 0.05& 0.07 & 0.10 &  0.04   & 0.12 & 0.04& 0.07  \\
 &       Random 1:1  & 0.23 & 0.11 &  0.04&0.04  & 0.04 & 0.09   &  0.05   & 0.10 & 0.07& 0.07  \\
  &    Minimal Mix-in  & 0.26 & 0.12 & 0.05 & 0.05 & 0.06& 0.11   &  0.05& 0.12 & 0.05 & 0.07\\
    \end{tabular}
    \caption{$\Delta$BLEU and $\Delta$COMET  on in-domain test sets for the same experiments as in Table \ref{tab:mixinbleu-generic}. Higher is better.
    }
    \label{tab:mixinbleu-indomain}
\end{table*}

In the previous section, we found that adaptation dataset vocabulary coverage has a strong negative correlation with forgetting. A natural question follows: what would be the effect of ensuring all of these adaptation datasets had 100\% vocabulary coverage?  To answer, we propose and perform \emph{Minimal Mix-in}, a targeted variant of mixed fine-tuning \cite{chu-etal-2017-empirical}. We focus on target coverage and the quality degradation metrics BLEU and COMET, both for brevity and as these had the strongest relationship with vocabulary coverage.

Mixed fine-tuning aims to mitigate forgetting by mixing examples from the generic training set into the adaptation set. For \emph{Minimal Mix-in}, we add generic examples to the adaptation set if they include a target token that is \emph{not in the adaptation dataset so far}. Aside from the novelty of targeted mix-in data, our goal is to examine the effect of improving the vocabulary coverage with minimal other change to the adaptation data and no change at all to the model architecture, adaptation or inference procedure. Excepting Kor and BSD, \emph{Minimal Mix-in} produces an adaptation dataset where fewer than 10\% of examples are generic.

To benchmark the  forgetting mitigation possible with a similar \emph{non}-minimal data augmentation,  we follow a popular mixed fine-tuning recipe found in the literature \cite{haque-etal-2020-terminology,hasler-etal-2021-improving}  which uses a 1:1 ratio of randomly sampled generic segments to in-domain  segments. We refer to this as \emph{Random 1:1}. Table \ref{tab:mixindatasets} summarizes the size of the different mix-in interventions.



\subsection{Less than 10\% of the mix-in data can mitigate  80\% of the forgetting}
In   Table \ref{tab:mixinbleu-generic} we verify that \emph{Random 1:1} generic data mix-in  does mitigate more forgetting than \emph{Minimal Mix-in}, or fine-tuning with no mix-in at all. However, \emph{Minimal Mix-in}  mitigates a large proportion of the forgetting, within 1 BLEU of \emph{Random 1:1} for 6 domains and within 0.03 COMET for 7. Assuming \emph{Random 1:1} benchmarks the forgetting mitigation possible while adjusting only data mix-in, the proportion of that mitigation achieved by \emph{Minimal Mix-in} is $\frac{No Mix\text{-}in - Minimal Mix\text{-}in}{No Mix\text{-}in - Random 1:1}$. This value is at least 80\% of the \emph{Random 1:1} forgetting mitigation for 6 domains when measuring $\Delta$BLEU and for 4 domains when measuring $\Delta$COMET -- and at least 70\% for 6 domains over both metrics. \emph{Minimal Mix-in} achieves this while mixing in less than 10\% as much generic data for all domains except Kor and BSD. It is worth noting that Sub and IWSLT, for which \emph{Minimal Mix-in} performs less well,  have high vocabulary coverage of the generic test set, as indicated in the discussion of Table \ref{tab:forgetting-metrics-split}. 

\emph{Minimal Mix-in} also reduces forgetting variation across the  domains. This is in line with our prior finding that vocabulary coverage for an in-domain adaptation set correlates strongly with forgetting. Our experiment effectively sets vocabulary coverage to be the same -- 100\% -- for every domain, which results in correspondingly very similar forgetting across all domains even if ensuring coverage does not mitigate forgetting entirely. This finding also supports work by Gu \& Feng \shortcite{gu-feng-2020-investigating} showing that, when adapting with frozen parameters, decoder embeddings are most correlated with  preserved generic performance.  Our results, from a data perspective,  suggest that future work might focus on specifically decoder embeddings for tokens not in the in-domain data.

Finally we examine vocabulary shift. We confirm by inspection that the less desirable replacements from Table \ref{tab:termshift-examples} are no more. For example, for the Med domain, $ForgetGenUse[\text{billion}]$ drops from 18 to 0, meaning everywhere the baseline model produces \emph{billion} correctly, so does the adapted model. Analysing $ForgetGenUse$, we find a pattern generally the same as for COMET and BLEU - \emph{Minimal Mix-in} is on par with a 1:1 generic ratio. The main exception is en-ja IWSLT. It is possible that domains where generic test vocabulary is almost entirely covered by the in-domain data already may benefit less from \emph{Minimal Mix-in}. As noted in Section \ref{sec:why},  $ForgetGenUse$ has less correlation with target coverage. A richer mix-in set may be required to address detrimental vocabulary shift.  

We note that a relationship between vocabulary shift and forgetting can be applied beneficially to intentional forgetting. In one brief experiment we adapted an English-to-German system on generic data with informal-you (\emph{du/ihr}, and inflections) target segments removed -- the resulting model only produced formal-you outputs.

\subsection{Minimal mix-in, better in-domain scores}

A primary goal of adapting NMT is improved in-domain translation.  Table \ref{tab:mixinbleu-indomain} gives quality metric deltas on the in-domain test sets: higher values are now better. A 1:1 generic  ratio has a negative impact, with noticeable BLEU and COMET drops relative to unmixed fine-tuning. By contrast,  \emph{Minimal Mix-in} scores similarly to unmixed fine-tuning for all except de-en Med. Improvement in terms of $\Delta$COMET shows less variation than under $\Delta$BLEU, possibly because COMET assigns higher scores to paraphrases which may not use domain-specific terminology.  Mixing in large amounts of generic data  reduces scores relative to \emph{Minimal Mix-in}. It is interesting to note that for the smallest domain, Kor, mixing no generic data also leads to reduced in-domain performance.

\section{Conclusions}
This paper investigates what is forgotten during NMT domain adaptation, and why. We show that vocabulary shift during adaptation is not necessarily beneficial, and that detrimental shift can be orthogonal to quality metrics. We find  forgetting correlates with in-domain vocabulary coverage, allowing better prediction of how adaptation will behave on a particular dataset. Our findings emphasise that NMT adaptation research should not be dataset agnostic: in-domain data characteristics are critical to how  adaptation can succeed or fail. 

\section*{Limitations}
Since our investigation is dataset-dependent, it is necessarily limited by the data and languages we have used. We report on a selection of widely used, diverse domain-specific datasets, as available for two language pairs with contrasting resources and distance. Additional language pairs or domains would allow us to generalise better.

Another limitation is model variety. In the interests of brevity, time and cost we only conduct our experiments with moderately sized Transformers trained for NMT. There has been much recent interest in machine translation by prompting Large Language Models (LLMs) pre-trained on huge uncurated datasets \cite{zhang-etal-2023-machine}. Work concurrent with ours by Pang et al \shortcite{salutetheclassic2024} observe that LLMs also struggle with domain-specific translation. Indeed, when fine-tuning on the same de-en OPUS domain-specific datasets as us, they report that LLMs exhibit similar behaviour in terms of `forgetting' domain-specific terminology in preference to tokens appearing in the adaptation set, although they do not attempt to explain or mitigate this. We leave confirming experiments to future work.



\bibliography{custom}

\appendix
\section{Experimental setup}
\label{appendix-setup}

We pre-train two Transformer models using the Tensorflow T2T toolkit \cite{vaswani-etal-2018-tensor2tensor}, one for each of German-English (de-en) and English-Japanese (en-ja). Both use BPE vocabulary  \cite{sennrich-etal-2016-neural}, with details given in Table \ref{tab:modelspecs}. Following findings from the most recent WMT shared task \cite{kocmi-etal-2023-findings} on Transformer NMT models, we use deep encoders with relatively shallow decoders for a balance of speed and quality. We found a slightly deeper encoder and smaller, not shared BPE vocabulary gave better results for en-ja in initial testing.

The de-en model is pre-trained on 43.9M lines of parallel data made available via the WMT shared task: Paracrawl v9, Europarl v10, NewsCommentary v14, Tilde and WikiMatrix \cite{kocmi-etal-2022-findings}. The en-ja model is pre-trained on 22.4M lines of JParacrawl v3.0 \cite{morishita-etal-2022-jparacrawl}.  When calculating BLEU for en-ja, we use Sacrebleu v2.0 \cite{post-2018-call} with the Mecab tokenizer. 

To minimize our computational and energy use, we pre-train each model only once on 4 GPUs for approximately two days. Each fine-tuning run of 20K steps takes approximately 1 additional hour of training.

\begin{table}[h]
    \centering
    \small
    \begin{tabular}{c|cc}
         & de-en &en-ja \\
         \hline
    Encoder layers  & 15 & 18\\
    Decoder layers  & 3 & 3 \\
    Hidden size     & 2560 &2560 \\
    Filter size     & 640 &640 \\
    \# BPE merges     & 32K  & 16K \\
    Shared BPE & Y & N\\
    \end{tabular}
    \caption{Pre-trained model specifications}
    \label{tab:modelspecs}
\end{table}

\section{Mixing in randomly-selected data to the same proportion as Minimal Mix-in}
\label{appendix-setup}
To expand on our experiments in section 4, we note that a large enough randomly selected dataset may cover the model vocabulary without requiring a more complicated sampling procedure. To investigate this, we randomly sample a set the same size as Minimal Mix-in for each domain: Random \#Minimal. 

Random \#Minimal mitigates forgetting similarly on average to Minimal Mix-in, despite their apparently different coverage statistics. The average target vocabulary coverage when including Minimal Mix-in is 0.97 across the five de-en domains and 0.95 across the three en-ja domains, while for Random \#Minimal average coverage is 0.65 for de-en and 0.70 for enja.  However, \emph{relative to the vocabulary in the generic test set,} the average target vocabulary coverage is 0.99 (both de-en and en-ja) for  Cover1,  and 0.98 (de-en) / 0.99 (en-ja) for Random \#Minimal. It is therefore less surprising that the two mix-in sampling methods result in similar generic test set forgetting. We note that, by attempting to cover all tokens, Minimal Mix-in incorporates outliers which may be rare in test sets. Selecting the same number of examples randomly may in fact be more generic-test-set relevant. 

If there is a known text on which we wish to minimize forgetting, it may even be preferable to mix-in data until high vocabulary coverage on that text, rather than aiming for global coverage. We evaluate this  in an oracle setting by comparing Minimal Test, which samples generic data until the generic test set vocabulary is covered, and Random \#Minimal Test, which samples the same number of examples randomly. The resulting generic mix-in proportion is less than 1\% for all but the smallest domains. Forgetting is increased  relative to Minimal Mix-in; there may be a minimum ratio of generic mix-in data below which benefit is limited. Nevertheless, forgetting with Minimal Test in terms of $\Delta$BLEU is still reduced relative to simple fine-tuning. Forgetting is also lower on average than  Random \#Minimal Test: targeted relevant-coverage sampling can outperform random sampling.

We also evaluate the impact on improved in-domain performance. Minimal Mix-in and Random \#Minimal give similar in-domain scores, which are also similar to the scores from Minimal Test / Random \#Minimal Test. This suggests that minimal size of the generic mix-in set may be more important than generic mix-in content for maintaining  in-domain improvement, and that there is a minimal mix-in size past which in-domain improvement does not increase.

\begin{table*}[h!]
    \centering
   \small 
    \begin{tabular}{p{1.8cm}c| ccccc|c| ccc|c}
    & &  \multicolumn{6}{c|}{de-en} &  \multicolumn{4}{c}{en-ja}\\
  &  Mix-in method     & IT & Kor &Law &Med  & Sub & Mean &  IWSLT  &KFTT &BSD & Mean \\
             \hline

  \multirow{4}{*}{$\Delta$BLEU}    &    Minimal Mix-in & 1.1 &4.1 & 1.3 & 1.7  &3.0 & 2.2 & 3.2  & 0.1 & 1.7& 1.7\\
&        Random \#Minimal &1.8 & 4.9  &1.6  &2.4 & 3.1 & 2.8  &2.5  & 0.0 & 1.4& 1.3\\
  & Minimal Test &2.0  & 6.5 & 2.9  & 3.2  & 5.0 &   3.9  & 4.5 & 1.0 & 2.0 & 2.5\\
   &    Random \#Minimal Test  &3.7  & 7.6 & 3.7 & 4.7& 5.2 & 5.0 & 4.7  & 1.5 &3.2 &3.1   \\
 \hline

  \multirow{4}{*}{$\Delta$COMET}   &     Minimal Mix-in & 0.04 & 0.09  & 0.03 & 0.04 & 0.04 & 0.05 & 0.03 & 0.01 & 0.04 & 0.03  \\
   &    Random \#Minimal & 0.03& 0.07 & 0.02 & 0.02 & 0.04 & 0.04  & 0.02& 0.01 &0.03 & 0.03  \\
   &    Minimal  Test &0.05 & 0.13 & 0.04& 0.05& 0.08 & 0.07 & 0.06  & 0.03&  0.03& 0.04 \\
&     Random \#Minimal Test & 0.07&0.15  & 0.06& 0.08& 0.08 &0.09 & 0.07  &0.04 & 0.09 & 0.07\\

\hline
    \multirow{4}{*}{$ForgetGenUse$}    & Minimal Mix-in & 0.05 & 0.07 & 0.04 & 0.05 & 0.06  & 0.05  &  0.12 & 0.08& 0.09& 0.10  \\
      &  Random \#Minimal &0.05 &0.07  & 0.04 & 0.05& 0.05 & 0.05 &   0.12 &  0.11 & 0.10 & 0.11 \\

   &     Minimal Test & 0.07&  0.09 & 0.02 & 0.08& 0.06 & 0.06 &   0.20 & 0.29 & 0.10&  0.20  \\
    &    Random \#Minimal Test &0.20 & 0.14 &0.06 &0.23 & 0.09 & 0.14 &  0.34 & 0.57 &0.14 &  0.35  \\

    \end{tabular}
    \caption{Forgetting metrics on generic test sets, varying the mix-in dataset when fine-tuning for 20K iterations in each case. Lower is better for all metrics. Negative scores indicate improvement.
    }
    \label{tab:mixinbleu-generic-appendix}
\end{table*}

\begin{table*}[h!]
    \centering
   \small 
       \begin{tabular}{cc| ccccc|c| ccc|c}
    &   &  \multicolumn{6}{c|}{de-en} &  \multicolumn{4}{c}{en-ja}\\
    & Mix-in method     & IT & Kor &Law &Med  & Sub & Mean & IWSLT  &KFTT &BSD & Mean \\

             \hline

    \multirow{4}{*}{$\Delta$BLEU}      & Minimal Mix-in & 7.9  &6.4&  5.6& 4.7 &3.4 &  5.6&3.6    &  10.3& 4.5& 6.1\\
      &  Random \#Minimal &8.2& 6.1 &5.6& 5.5&  3.4 &  5.8& 3.6 &  10.4&4.6 & 6.2\\

  & Minimal Test & 8.4 &5.9 & 5.7 & 6.1 &  3.4&   5.9&3.7 & 10.5 & 4.7& 6.3\\
   &    Random \#Minimal Test  & 8.3  &  5.5 &  5.7 & 6.1 & 3.5 &   5.8&  3.6 &   10.4 & 4.4& 6.1  \\
 \hline        
 \multirow{4}{*}{$\Delta$COMET}  &    Minimal Mix-in  & 0.26 & 0.12 & 0.05 & 0.05 & 0.06& 0.11   &  0.05& 0.12 & 0.05 & 0.07\\
      &  Random \#Minimal  & 0.26 & 0.11 & 0.05  &  0.05 &0.07 &0.11  &0.05  & 0.12 & 0.07& 0.08 \\
  & Minimal Test  & 0.27 & 0.10 &  0.05& 0.05 &0.07 & 0.11  &0.05  &0.12  & 0.06& 0.08  \\
  &  Random \#Minimal Test  & 0.26 & 0.10 & 0.05  & 0.05 & 0.07 & 0.11  & 0.04  & 0.12 & 0.06&  0.07 \\

    \end{tabular}
    \caption{$\Delta$BLEU and $\Delta$COMET  on in-domain test sets for the same experiments as in Table \ref{tab:mixinbleu-generic}. Higher is better.
    }
    \label{tab:mixinbleu-indomain-appendix}
\end{table*}

\end{document}